\documentclass[11pt,x11names]{article}

\usepackage{setspace}
\usepackage[]{misc/acl}

\usepackage{times}
\usepackage[table,x11names]{xcolor}
\usepackage{graphicx}
\usepackage[T1]{fontenc}
\usepackage[utf8]{inputenc}
\usepackage{microtype}
\usepackage{longtable}
\usepackage{booktabs}
\usepackage{todonotes}
\usepackage{amsmath,amssymb}
\usepackage{cleveref}
\usepackage{xspace}
\usepackage{soul} % for \ul and \TODOMARK
\usepackage{caption}
\usepackage{multirow,multicol}
\usepackage{ulem}
\usepackage{float}
\usepackage{array}
\usepackage{bm}
\usepackage{xfrac}
\usepackage{tcolorbox} % for frames
\usepackage[shortlabels]{enumitem}
\usepackage[outline]{contour}% http://ctan.org/pkg/contour
\usepackage{tikz}
\usetikzlibrary{decorations.text,calc,arrows.meta}
\usepackage{hyperref}
\usepackage{pgfplots}
\usepackage{lipsum}
\usepackage{expex}
\usepackage{relsize}

\usepackage[safe]{tipa}

\pgfplotsset{compat=1.18}

\setlist[itemize]{noitemsep,left=0mm}

\usepackage{tabularx}

\crefname{lstlisting}{listing}{listings}
\Crefname{lstlisting}{Listing}{Listings}
\crefname{equ}{equation}{equations}
\Crefname{equ}{Equation}{Equations}
\Crefname{algorithm}{Algorithm}{Algorithms}
\crefname{example}{example}{examples}
\Crefname{example}{Example}{Examples}
\crefname{prompt}{prompt}{prompts}
\Crefname{prompt}{Prompt}{Prompts}
\DeclareCaptionType{example}[Example][List of examples]
\DeclareCaptionType{prompt}[Prompt][List of prompts]
\DeclareCaptionType{equ}[Equation][List of equations]

\usepackage{marginnote}

\usepackage{ifthen}
\newboolean{WIP}

\setboolean{WIP}{false}

\ifthenelse{\boolean{WIP}}{
    \definecolor{TodoColor}{rgb}{1,0.8,0.1} %yellow
    \definecolor{TodoColor2}{rgb}{1,0.3,0.9} %blue
    \definecolor{TodoColor3}{rgb}{0,0.9,0.6} %green
    \newcommand{\todonote}[3][]{\todo[color=#2,size=\scriptsize,fancyline,caption={},#1]{#3}}
    \newcommand{\todox}[2][]{\todonote[#1]{TodoColor}{\textbf{TODO:} #2}}
    \newcommand{\sa}[2][]{\todonote[#1]{orange!30}{\textbf{Seth:} #2}}
    \newcommand{\SA}[2][]{\sa[inline,#1]{#2}}
    \newcommand{\tm}[2][]{\todonote[#1]{blue!30}{\textbf{Timo:} #2}}
    \newcommand{\TM}[2][]{\TM[inline,#1]{#2}}
    
    \newcommand{\TODOBOX}[2][]{\todox[inline,#1]{#2}}

}{
    \newcommand{\todonote}[3][]{}
    \newcommand{\todox}[2][]{}
    \newcommand{\sa}[2][]{}
    \newcommand{\SA}[2][]{}
    \newcommand{\tm}[2][]{}
    \newcommand{\TM}[2][]{}
    \newcommand{\TODOBOX}[2][]{}

}

\makeatletter\def\Hy@Warning#1{}\makeatother
\let\svthefootnote\thefootnote
\newcommand\blankfootnote[1]{%
  \let\thefootnote\relax\footnotetext{#1}%
  \let\thefootnote\svthefootnote%
}

\graphicspath{{img/}}
\title{
    Typological Feature Induction as a Benchmark for\\Inductive Metalinguistic Reasoning
}

\author{
Chen Zhang$^{1}$\thanks{Equal contribution.} \quad Seth Aycock$^{2}$\footnotemark[1] \quad Yusuke Miyao$^{1,3}$ \\
$^{1}$NII LLMC \quad $^{2}$University of Amsterdam \quad $^{3}$The University of Tokyo \\
\texttt{\href{mailto:zhangchen@nii.ac.jp}{zhangchen@nii.ac.jp}} \quad
\texttt{\href{mailto:s.aycock@uva.nl}{s.aycock@uva.nl}} \\
\texttt{\href{mailto:yusuke@is.s.u-tokyo.ac.jp}{yusuke@is.s.u-tokyo.ac.jp}}
}
\begin{document}

\maketitle

\maketitle

\begin{abstract}
\label{sec:abs}
Existing metalinguistic reasoning benchmarks require both inference of rules and their application, making failures hard to diagnose. We introduce typological feature induction (TFI) to target the former step. Given sentences in an unnamed language paired with translations in a known pivot language, models must predict a typological feature of the unnamed language. From Grambank and multi-parallel corpora, we automatically build evaluation questions over 195 languages. We find that LLMs identify 59--80\% of these languages from only 16 sentence pairs; and misaligning the translations barely affects their accuracy, suggesting reliance on prior knowledge of the language. Enciphering text to prevent recognition increases the corpus' utility, but mainly for features where target and pivot languages agree. On languages which models cannot identify, neither open model outperforms a majority-value baseline. Since induction from raw text aligns with field linguists' use-case, this work indicates the models tested are not yet ready for this role. TFI offers a scalable path to track progress in this direction\footnote{The code and data are available \href{https://github.com/luciusssss/tfi}{here}.}.

\end{abstract}

\section{Introduction}
\label{sec:int}

Metalinguistics is the study of language phenomena using a \textit{metalanguage}~\citep{schneider-anastasopoulos-2026-speaking}, either as natural language \textit{about} language, or symbolic notation \textit{encoding} language information. While central to field linguistics, language acquisition and language teaching, questions regarding metalinguistic abilities of NLP models remain underexplored. Of particular interest is metalinguistic \textit{reasoning}: the task of applying reasoning methods to solve problems \textit{about} language.

Recent metalinguistic reasoning work evaluates LLMs on tasks such as rosetta stone-style linguistic Olympiad problems~\citep{sanchez2026linguini, bean2024lingoly}, translation from a grammar book~\citep{Aycock24-CanLLMsReallya, Tanzer24-BenchmarkLearningTranslatea}, or interlinear gloss prediction from grammar snippets~\citep{yang-etal-2025-linggym}.
Such tasks remain difficult for frontier models~\citep{choudhary2025unveiling, marmonier-etal-2025-explicit}.
We posit that these tasks' difficulty arises partly from conflating different aspects of metalinguistic reasoning: inductive reasoning (hypothesising rules of the linguistic system), and analogical reasoning (applying the system to new data).

\begin{figure}[t]
\centering
\includegraphics[scale=0.9]{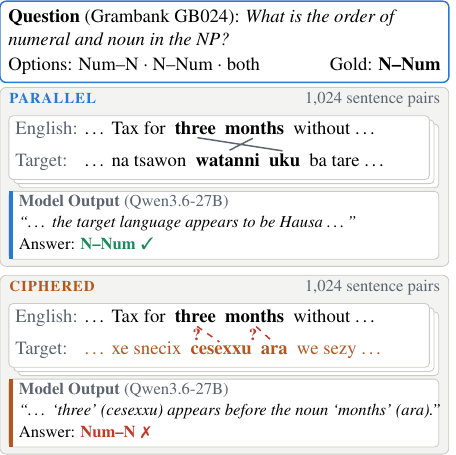}
\caption{A TFI item for Hausa, with an example sentence pair, and reasoning excerpts from Qwen3.6. In \textsc{parallel}, the model names the language and answers correctly by recall. With the target side enciphered, it aligns \textit{three months} word by word with \textit{watanni uku} (`months three') and answers with the English order.}
\label{fig:teaser}
\end{figure}

In light of this division, we propose typological feature induction (TFI), a task that isolates LLMs' inductive metalinguistic reasoning based on raw data, while also probing understanding of typology, a fundamental facet of linguistic analysis \cite{ploeger-etal-2026-principled}.
A model is shown $n$ sentence pairs between a known pivot language and an unnamed target language, and is asked for the value of one typological feature of the target, such as the order of numeral and noun (Figure~\ref{fig:teaser}). Questions and gold values come from Grambank~\citep{Skirgard23-GrambankV1} and the sentences from multi-parallel corpora, so items are built automatically for any language covered by both, and each answer is a single property that can be checked against expert coding.

TFI also exposes the central difficulty of evaluating induction in LLMs. These models have seen many of the world's languages, and withholding the name does not hide the language: from 16 sentence pairs, the three models we test name 59--80\% of our 195 target languages. A correct answer may therefore be recalled rather than induced. We pursue three research questions:

\begin{enumerate}[label=\textbf{RQ\arabic*}, itemsep=0pt, leftmargin=*,topsep=2pt, partopsep=0pt]
    \item To what extent can LLMs infer typological features from a parallel corpus? (\S\ref{sec:res-saturate})
    \item How much of this reflects language recall rather than induction from the corpus? (\S\ref{sec:res-langid})
    \item When recall is removed, what do the models take from the corpus? (\S\ref{sec:res-cipher}--\S\ref{sec:dis-extended})
\end{enumerate}

To answer them, we vary what the corpus contains: we remove the translations or break the sentence alignment, encipher the target side so that the language cannot be recognised, and replace English with 14 other pivot languages.

In plaintext, a corpus raises accuracy, but the gain saturates within 16 pairs, and removing the translations, breaking the alignment or adding the language name barely changes it (\S\ref{sec:res-saturate}). The models name the language in their reasoning and answer from knowledge of it (\S\ref{sec:res-langid}). 
Once the cipher makes the language unrecognisable, more pairs, the translations and their alignment all help (\S\ref{sec:res-cipher}), but this help comes from the pivot: it raises accuracy only where the target shares the pivot's value, and pulls answers towards that value (\S\ref{sec:res-pivot}). Finally, on languages none of the models can identify, neither open model beats a majority baseline across Grambank's features (\S\ref{sec:dis-extended}).

\noindent Our contributions are as follows:
\begin{itemize}[itemsep=0pt,topsep=2pt, partopsep=0pt]
    \item We introduce TFI, which isolates inductive metalinguistic reasoning from rule application, with a benchmark generated from Grambank covering 195 languages.
    \item Through controlled corpus ablations, pivot swaps, and ciphering, we show that corpus gains are largely attributable to recall.
    \item We identify two shortcuts that look like induction: recognising the language and recalling its grammar, and projecting the pivot language's grammar onto the target. 
    \item With both shortcuts removed, the models do not beat a majority baseline across Grambank's features, indicating substantial headroom for inductive metalinguistic reasoning.
\end{itemize}

\section{Methodology}
\label{sec:met}

\subsection{Isolating Induction from Analogy}

\paragraph{Theoretical Framework}
We introduce the task of typological feature induction (TFI): given $n$ unlabelled sentence-level data points $D$ in target language $L_t$ and paired translations in pivot language $L_p$ (English unless specified), the task is for model $\theta$ to predict values $v_{1..i}$ for typological features $F_{1..i}$.
Figure~\ref{fig:teaser} exemplifies the task, defined as:
\vspace{-0.1cm}
\begin{equation}
    \hat{v}_{1..i} = f_\theta(D^{L_{t}, L_{p}}_{n}, F_{1..i}), \qquad \hat{v}_j \in \mathcal{V}_j
    % \hat{v}_{1..i} = f_\theta(D, F_{1..i}), \qquad \hat{v}_j \in \mathcal{V}_j
\end{equation}
where $\mathcal{V}_j$ is the value set for feature $F_j$ (e.g. $\{0,1\}$ or $\{0,1,2\}$ in Grambank). Evaluation is exact match accuracy, comparing $\hat{v}_i$ against the gold value $v_i^{\text{gold}}$ for $L_t$. Under~\citeposs{schneider-anastasopoulos-2026-speaking} taxonomy, TFI instantiates a system-level, multilingual, processing-focused, natural metalinguistic task. Our characterisation of TFI focuses on LLMs' inductive understanding rather than analogical (and generative) capabilities.

\paragraph{Data Sources}
We source features $F$ and values $v$ from Grambank~\citep{Skirgard23-GrambankV1}, a database of 195 typological feature specifications in 2,467 languages. Grambank defines features in terms of questions, including a short linguistic description and possible values across languages.

Unlabelled language data $D$ is derived from FLORES+~\citep{NLLBTeam22-NoLanguageLeft} and BOUQuET~\citep{andrews-etal-2025-bouquet}, both multi-parallel human-translated sentence-level corpora. We select languages in Grambank matched through Glottolog~\citep{glottolog5_3}, resulting in 195 target languages  (See Appendix~\ref{app:lang-list} for full lists).

\subsection{Isolating Induction from Recall}
\label{sec:met-controls}

A model that has seen a language in pretraining can answer from memory rather than from the corpus; that is, the model relies on parametric \textit{recall} rather than \textit{induction}. 
We first probe how often the text gives the language away.
We then vary the input in three ways (Table~\ref{tab:settings}): ablations remove part of the corpus, a cipher hides the language's identity, and a pivot swap changes the known side. 

\paragraph{Language Identification} To measure how often the text gives the language away, we show each model $n \in \{16, 64, 256\}$ unlabelled pairs and ask it to name the target language. A guess counts as correct if it gives the language's name or a recognised variant of it.

\paragraph{Plaintext Ablations} \textsc{mono} removes the pivot side and \textsc{misaligned} shuffles the target-side lines so that pairs no longer match; the model is not told that the pairs are misaligned. \textsc{parallel+name} adds the language name to \textsc{parallel}.

\paragraph{Cipher} To remove the shortcut through recognition, we encipher the target side, following prior work that uses substitution ciphers to test reasoning without memorisation~\citep{khouja2026lingolytoo}. Permuting characters within a language's own alphabet is not enough: 87\% of the Latin-script languages have a unique set of non-ASCII characters (\textit{ą ć ę ł} is Polish), and under such a cipher Qwen3.6 still named 3--9 languages per run, citing the diacritics in its reasoning. Our cipher therefore decomposes each character into a base letter and a modifier and maps the two separately, drawing the targets from a pool shared across all languages: if \textit{a} is mapped to \textit{e}, then \textit{á} is mapped to \textit{è}. A language's diacritic inventory is thereby replaced rather than permuted. Vowels map to vowels and consonants to consonants, so the text stays plausible as a natural language. 
Its cost is that enciphered text tokenizes approximately 1.3$\times$ longer, which is affordable. 

\paragraph{Pivot Swap} To test what the models take from the known side, we replace English in \textsc{cipher-parallel} with 14 other pivots, chosen to cross transitive verb position (verb-medial, -final, -initial) with script (Latin, non-Latin): French, Indonesian, Swahili, Russian, Hebrew, Mandarin, Turkish, Hungarian, Korean, Hindi, Persian, Japanese, Welsh and Egyptian Arabic. 
A pivot that is also a target is removed from the targets in its own runs.

\section{Experimental Setup}
\label{sec:met-eval}

\paragraph{Baselines} \textsc{name-only} gives the language name and no corpus to measure pure recall. \textsc{majority} answers every question with the most frequent value across all 2,467 Grambank languages, and \textsc{random} is the expected accuracy of a uniform guess over the options.

\paragraph{Models}
We evaluate two open-weight models, Gemma~4~31B-it~\citep{gemmateam2026gemma4} and Qwen3.6-27B~\citep{qwen3.6-27b}, and one closed model, GPT-5.6 Luna~\citep{gpt5.6}. 
For each language we sample corpora of $n \in \{16, 32, 64, 128, 256, 512, 1024\}$ sentence pairs. Corpora are nested: the 16 pairs at $n{=}16$ are the first 16 at $n{=}32$ etc. Each open-model setting is run with three corpus seeds; GPT-5.6 Luna is run once per setting for cost reasons.
See implementation details and prompt templates in Appendix~\ref{app:impl}.

\begin{table}[t]
\centering\small
\setlength\tabcolsep{3.5pt}
\begin{tabular}{ll}
\toprule
 & Description \\
\midrule
\textsc{parallel} & $n$ aligned pairs \\
\textsc{mono} & $n$ target sentences only \\
\textsc{misaligned} & $n$ pairs, target side shuffled \\
\textsc{parallel+name} & $n$ aligned pairs $+$ language name \\
\textsc{cipher-parallel} & $n$ pairs, target side enciphered \\
\textsc{cipher-mono} & \textsc{mono} on enciphered text \\
\textsc{cipher-misaligned} & \textsc{misaligned} on enciphered text \\
\midrule
\textsc{name-only} & language name, no corpus \\
\textsc{majority} & most frequent value in Grambank \\
\textsc{random} & random guess over options \\
\bottomrule
\end{tabular}
\caption{Experimental settings and baselines. }
\label{tab:settings}
\end{table}

\paragraph{Evaluation Sets} We evaluate on three sets:
\begin{itemize}[itemsep=4pt,topsep=2pt, partopsep=0pt]
    \item \textbf{Core:} We manually select 22 features deemed to be inferrable from parallel sentences, covering word order, word classes and morphology. Every Grambank-coded pair of these features with the 195 languages gives 4,084 items. See the list in Appendix~\ref{app:features}.
    \item \textbf{Latin-Script:} We only apply the cipher (\S\ref{sec:met-controls}) to Latin-script orthographies because non-Latin scripts are more likely to reveal the language and might require completely different ciphering methods. We obtain a 2,657-item subset for 127 languages.
    % Of the 134 Latin-script languages, 127 can be enciphered; the remaining seven use IPA letters or more vowel graphemes than the cipher's character pool holds. Their items in the core set form the 2,657-item subset.
    \item \textbf{Extended:}  This set tests features beyond the 22 core ones. Asking about every coded feature for all 195 languages would give 32,040 items, which is too costly to run. We therefore restrict this set to the languages that none of the models can identify. Here a language counts as identified if a model names it, or the broader language it belongs to, in more than half of its language-identification runs. For these 24 languages we ask about every feature Grambank codes, giving 3,775 items over all 195 features.
\end{itemize}

\begin{figure*}[t]
\centering
\includegraphics{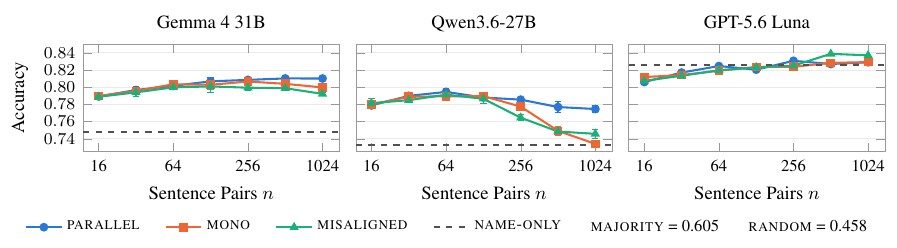} 
\caption{Accuracy of TFI on the core evaluation set as the number of sentence pairs $n$ grows. Error bars show the standard deviation over three corpus seeds for the open models; GPT-5.6 Luna was run on one.}
\label{fig:saturation}
\end{figure*}

\section{Results}
\label{sec:res}
We first measure how much a plaintext corpus helps (RQ1, \S\ref{sec:res-saturate}) and how much of that help comes from recognising the language (RQ2, \S\ref{sec:res-langid}). We then block recall in two ways and ask what the models take from the corpus instead (RQ3): by enciphering the target side (\S\ref{sec:res-cipher}), which also lets us swap the pivot language (\S\ref{sec:res-pivot}), and by choosing languages the models cannot identify (\S\ref{sec:dis-extended}).

\subsection{Corpus Effects}
\label{sec:res-saturate}
To answer RQ1, we start from plaintext corpora. Figure~\ref{fig:saturation} gives accuracy on the core evaluation set. Both open models answer far above \textsc{majority} (0.605) even without a corpus, and a corpus raises their accuracy further: with 128 pairs, Gemma~4 gains 5.9 points over \textsc{name-only} and Qwen3.6 5.6 points.
GPT-5.6 Luna starts higher (0.826) and gains nothing: no corpus setting is more than about one point above \textsc{name-only}, and with 16 pairs it is two points below it.

\paragraph{Few pairs give strong performance.}
For both open models, 16 pairs already deliver 70--85\% of the gain observed at 128 pairs. Beyond that the curves flatten. From 16 to 128 pairs Gemma~4 gains 1.8 points and Qwen3.6 0.9 points. From 128 to 1,024 pairs Gemma~4 gains 0.3 points, and Qwen3.6 \emph{loses} 1.3 points. Qwen's decline is almost entirely a failure to answer rather than wrong answers: the number of unparsed outputs rises from about 5 to about 51 per run between 128 and 1,024 pairs (Appendix~\ref{app:unparsed}).
GPT-5.6 Luna shows the same shape at a higher level, rising from 0.806 at 16 pairs to 0.829 at 1,024, no better than \textsc{name-only}.

\paragraph{Parallelism and alignment barely matter.}
If the models induced features by aligning target sentences with their translations, removing the English side (\textsc{mono}) or breaking the alignment (\textsc{misaligned}) should hurt. Up to 128 pairs neither does: at all four corpus sizes from 16 to 128 pairs, \textsc{parallel} differs from \textsc{mono} and from \textsc{misaligned} by at most 0.6 points for either open model, with no consistent sign (Figure~\ref{fig:saturation}). 
The same holds for GPT-5.6 Luna, whose \textsc{mono} score (0.835) is at least as high as its \textsc{parallel} score. 

Differences appear only with long corpora and remain small. For Gemma~4, alignment is worth 0.9, 1.1 and 1.8 points at 256, 512 and 1,024 pairs, and parallelism 1.0 point at 1,024. Qwen's larger gaps at these sizes (up to 4.1 points at 1,024) come mostly from unanswered items: its \textsc{mono} and \textsc{misaligned} runs fail to produce an answer for 60--206 items per run, against 17--51 for \textsc{parallel}.
Most of these outputs do not loop verbatim but keep going through corpus sentences without reaching a conclusion (Appendix~\ref{app:unparsed}).

\begin{table}[t]
\centering\footnotesize
\begin{tabular}{@{}>{\raggedright\arraybackslash}p{0.97\linewidth}@{}}
\toprule
\textbf{(a) Language naming.} Qwen3.6, \textsc{parallel}, $n{=}128$; Estonian, GB075 (postpositions); correct. \\
\addlinespace[2pt]
``Based on the provided corpus, the target language (\textbf{Estonian}) exhibits a strong preference for postpositions [\dots] \textbf{In Estonian}, many spatial and temporal relationships are expressed using nouns in the genitive case followed by a postpositional noun [\dots]'' \\
\addlinespace[2pt]
In \textsc{cipher-parallel} the same model names no language and reaches the same answer after 18,847 characters of analysis (2,620 in plaintext). \\
\midrule
\textbf{(b) Corpus citation.} Gemma~4, \textsc{misaligned}, $n{=}128$; Mískito, GB044 (bound plural marking); correct. \\
\addlinespace[2pt]
``Reviewing the corpus: [\dots] 5.~`issues' $\rightarrow$ \textit{niks munia}. [\dots]'' \\
\addlinespace[2pt]
The target line shown with the \textit{issues} sentence translates a recipe (\textit{Diara mangki \textbf{niks munia} dukia nani: 1 kg tawa, \dots}; `The ingredients needed include: 1 kg potatoes, \dots'), so the gloss is invented. In \textsc{parallel}, Gemma~4 glosses \textit{issues} as \textit{aisan nani}. \\
\bottomrule
\end{tabular}
\caption{Two reasoning patterns: (a) Naming the language and answering from recall. (b) Citing corpus pairs as evidence, including pairs that do not match.}
\label{tab:reasoning-examples}
\end{table}

\paragraph{Models cite the corpus, regardless of alignment.}
Most outputs quote the corpus. On 1,000 random items at $n{=}128$, 79\% of Gemma~4's outputs, 86\% of Qwen3.6's and 92\% of GPT-5.6 Luna's quote at least one corpus sentence, on average 3.4, 5.6 and 3.5 of the 128. They quote as often in \textsc{misaligned} (80\%, 89\% and 94\%).
Gemma~4 and GPT-5.6 Luna rarely say that the pairs do not match. We label the same 1,000 outputs using an LLM judge. 3.5\% of Gemma~4's \textsc{misaligned} outputs and 0.2\% of GPT-5.6 Luna's say so. Qwen3.6 says so in 33\%, and about half of these say that the whole corpus is shuffled, yet its accuracy is the same as in \textsc{parallel}. In Table~\ref{tab:reasoning-examples}b, Gemma~4 even glosses an English word with target words from an unrelated sentence. Therefore a quoted sentence does not reliably indicate that the answer came from it.

\paragraph{Language name adds nothing beyond corpora.}
Giving the language name on top of the corpus (\textsc{parallel+name}) leaves Gemma~4 unchanged at 128 pairs ($+$0.06 points) and lowers Qwen3.6 by 1.0 point. Whatever information the name carries, the target-language text already supplies it.

\paragraph{Interpretation}
Taken together, the corpus behaves less like evidence to be analysed than like a cue: a handful of target-language sentences is enough to reach the plateau, the translations and their alignment add little, naming the language adds nothing further, and the pairs the models cite need not match. This is the pattern expected if the text mainly activates what the model already knows about the language, which we test next.

\subsection{Language Identification}
\label{sec:res-langid}

\begin{table}[t]
\centering\footnotesize
\setlength\tabcolsep{3.5pt}
\begin{tabular}{llccc}
\toprule
 & & $n{=}16$ & $n{=}64$ & $n{=}256$ \\
\midrule
\multirow{3}{*}{\shortstack[l]{before cipher\\(all, 195)}} & Gemma~4 & 0.660 & 0.662 & 0.658 \\
 & Qwen3.6 & 0.593 & 0.581 & 0.576 \\
 & GPT-5.6 Luna & 0.800 & 0.809 & 0.805 \\
\midrule
\multirow{3}{*}{\shortstack[l]{before cipher\\(Latin, 127)}} & Gemma~4 & 0.643 & 0.646 & 0.640 \\
 & Qwen3.6 & 0.583 & 0.562 & 0.559 \\
 & GPT-5.6 Luna & 0.795 & 0.811 & 0.806 \\
\midrule
\multirow{3}{*}{\shortstack[l]{after cipher\\(Latin, 127)}} & Gemma~4 & 0.000 & 0.000 & 0.000 \\
 & Qwen3.6 & 0.008 & 0.000 & 0.000 \\
 & GPT-5.6 Luna & 0.039 & 0.039 & 0.031 \\
\bottomrule
\end{tabular}
\caption{Language identification accuracy, i.e. the proportion of languages each model names correctly from $n$ unlabelled sentence pairs.}
\label{tab:langid}
\end{table}

\begin{figure}[t]
\centering
\includegraphics{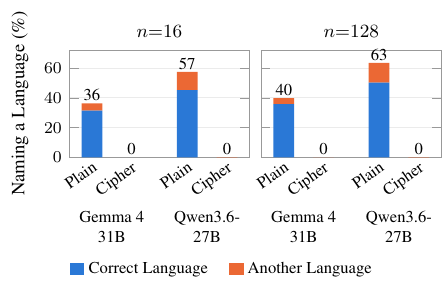} 
\caption{Proportions of outputs that name a specific language, in \textsc{parallel} and \textsc{cipher-parallel}, on the Latin-script evaluation set.}
\label{fig:reasoning}
\end{figure}

\paragraph{Models can identify the language.}
For RQ2, we first ask whether the models recognise the language at all. We give each model $n$ sentence pairs and ask it to name the target language (Table~\ref{tab:langid}). From only 16 pairs, Gemma~4 names 66\% of the 195 languages correctly, Qwen3.6 59\% and GPT-5.6 Luna 80\%; more pairs do not help. 
On the 127 Latin-script languages to which the cipher applies, the rates are almost the same (64\%, 58\% and 80\%); these languages are no less recognisable than the rest. 
This indicates that the LLMs have substantial exposure to the target languages during pretraining.

\paragraph{Models name languages while answering.}
To see whether the models also name the language while answering the typological questions, we use gpt-oss-20b~\citep{openai2025gptoss120bgptoss20bmodel} as a judge on the outputs of the two open models in \textsc{parallel} and \textsc{cipher-parallel}, using the Latin-script evaluation set. The judge extracts the language that the output identifies as the target (Appendix~\ref{app:langname}).

In \textsc{parallel}, Gemma~4 names a specific language in 36--40\% of its outputs and Qwen3.6 in 57--63\%, mostly the correct one (Figure~\ref{fig:reasoning}). The name usually comes in the first sentence, and the output then draws on knowledge of that language (Table~\ref{tab:reasoning-examples}a). Both models name the language more often with more corpus: from $n{=}16$ to $n{=}128$ the share rises by 3 points for Gemma~4 and 6 points for Qwen3.6, presumably because more sentences give more cues for recognising the language.

\paragraph{Identification is not the whole story.}
The gain is not confined to languages the model can name: for the languages Gemma~4 cannot identify from 16 pairs, the corpus adds 7.1 points over \textsc{name-only} at 128 pairs, more than the 5.3 points it adds for the languages it names. The text may therefore also reveal partial knowledge, such as the language's family or area. Either way, plaintext accuracy mixes recall with whatever the models induce, so we next remove the language's identity from the text.

\providecommand{\std}[1]{{\scriptsize$\pm$#1}}
\begin{figure}[t]
\centering
\includegraphics{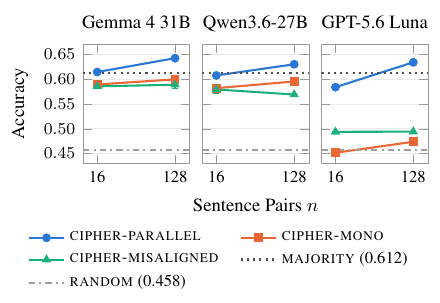}  
\caption{Accuracy of TFI on the 127 Latin-script languages with the target side enciphered. Error bars show the standard deviation over three corpus seeds.}
\label{fig:cipher}
\end{figure}

\subsection{Induction and Ciphering}
\label{sec:res-cipher}
For RQ3, we first block recall by enciphering the target side and ask whether the corpus matters.

\paragraph{Ciphers provide a clean testbed.}
The cipher all but removes identification on the 127 languages (Table~\ref{tab:langid}, bottom): from 64\% to 0\% for Gemma~4 and from 58\% to at most 0.8\% for Qwen3.6. GPT-5.6 Luna still names 3--4\%, all from a handful of well-known European languages (Italian at every $n$). 
Nor do the models recognise the language while answering TFI items: under the cipher almost no output names a language (Figure~\ref{fig:reasoning}). The cipher thus largely blocks the route through recall.

\paragraph{After ciphering, corpus size, parallelism and alignment all matter.}
Figure~\ref{fig:cipher} shows accuracy on the 127 enciphered languages. Unlike in plaintext, what the input contains now makes a clear difference. Going from 16 to 128 pairs raises accuracy by 2.8 points for Gemma~4 and 2.3 for Qwen3.6. Without the pivot side (\textsc{cipher-mono}), Gemma~4 loses 2.5 points at 16 pairs and 4.3 at 128, and Qwen3.6 loses 2.6 and 3.5. Shuffling the pairs (\textsc{cipher-misaligned}) costs more still: 2.9 and 5.3 points for Gemma~4, 2.8 and 6.1 for Qwen3.6.
Once the models can no longer recognise the language, they do make use of the English side and of how it lines up with the target.

\subsection{Pivot Language Effects}
\label{sec:res-pivot}

\begin{figure*}[t]
\centering
\includegraphics{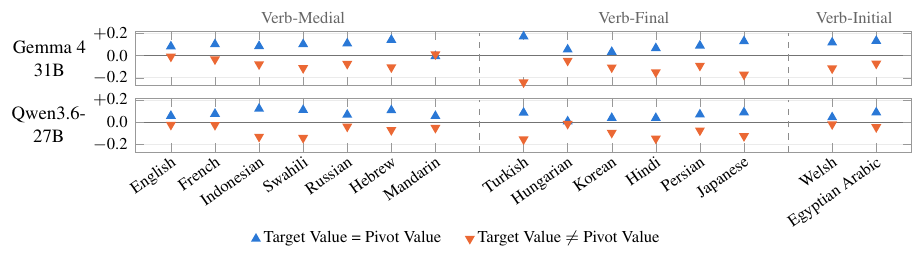} 
\caption{Change in accuracy from adding different pivot sides to an enciphered corpus ($n{=}128$; \textsc{cipher-parallel} $-$ \textsc{cipher-mono}).}
\label{fig:pivot}
\end{figure*}

What exactly do the models take from the pivot side (RQ3)? If they used it to locate constructions in the target, the benefit should not depend on what the pivot's own grammar looks like. 
We run the pivot swap (\S\ref{sec:met-controls}) on the ciphered data at $n{=}128$ with 15 pivots including English.

\paragraph{Pivots help only where their grammar is the answer.}
For each item we ask whether the target's Grambank value equals the pivot language's own value for that feature. Figure~\ref{fig:pivot} shows the gain over \textsc{cipher-mono} for the two kinds of item. The pattern is the same for almost every pivot: 
adding the pivot side raises accuracy where the target agrees with the pivot and lowers it where it does not. 
The models appear to read the target through the pivot's grammar rather than through the target's own: with Japanese (verb-final) as the pivot, Gemma 4 calls 74\% of the non-verb-final targets verb-final, against 15\% without a pivot. Figure~\ref{fig:teaser} shows the same projection on a single item: Qwen3.6 aligns an enciphered noun phrase with its English translation word by word and reads off the English order.

\begin{figure}[t]
\centering
\includegraphics{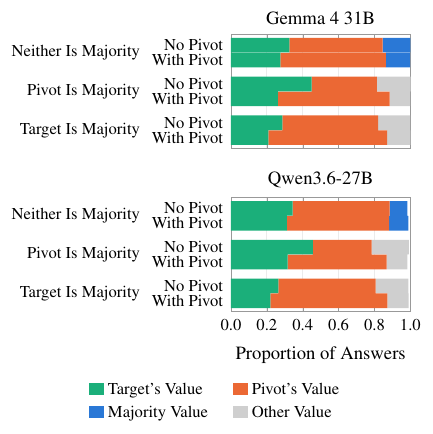}
\caption{Distribution of answers on Grambank features, for items where the target's value differs from the pivot's, without the pivot side (\textsc{cipher-mono}) and with it (\textsc{cipher-parallel}).}
\label{fig:pivot-choice}
\end{figure}

\paragraph{Pivots pull answers to their own values.}
A pivot that agrees with the target raises accuracy, but this has two explanations: the models may use the pivot to analyse the target, or they may simply copy the pivot's value, which is then also correct. Items where target and pivot disagree tell the two apart. If the models analysed the target, the pivot side should make them give the target's value more often. Instead, it makes them give the pivot's value more often (Figure~\ref{fig:pivot-choice}). Over all items, adding the pivot side raises the chance of answering with the pivot's value by 20 points for Gemma~4 and 14 for Qwen3.6, and does not raise the chance of answering with the target's value (Appendix~\ref{app:pivot}). The pull is strongest on word order (Figure~\ref{fig:pivot-feature} in Appendix~\ref{app:pivot}).

This qualifies the result of \S\ref{sec:res-cipher}. Once the target is unrecognisable, the parallel side does change the answers, but mostly by moving them towards the pivot's values. With English as the pivot, this projection looks like induction, because English shares its value with the target on 62\% of items.

\subsection{Extending the Feature Set}
\label{sec:dis-extended}

\begin{figure}[t]
\centering
\includegraphics{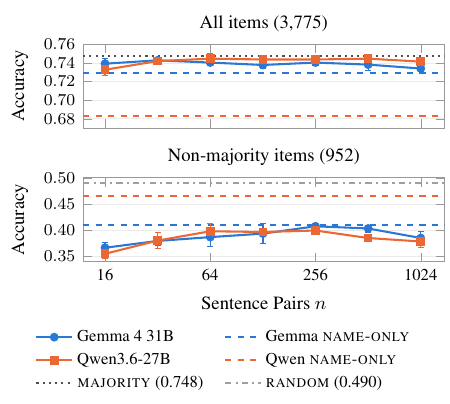}  % 原尺寸插入；源文件 figures/extended.tex
\caption{Accuracy on the extended set as the number of sentence pairs $n$ grows. The top panel shows all 3,775 items, the bottom panel the 952 items whose answer is not the most frequent value in Grambank. Error bars show the standard deviation over three corpus seeds.}
\label{fig:extended}
\end{figure}

Another way to block recall for RQ3 is to choose languages the models do not know. The extended set does this, and by asking about every Grambank feature coded for these languages it also tests whether a model can find a less common value.

\paragraph{Models do not beat the majority baseline.}
On all items, neither model reaches \textsc{majority} (0.748) at any corpus size (Figure~\ref{fig:extended}, top). \textsc{name-only}, which shows what the models recall about these languages, is already below \textsc{majority} (0.729 for Gemma~4, 0.683 for Qwen3.6), and the corpus brings both close to \textsc{majority} but not above it. 

The non-majority items, where the most frequent value is wrong, show whether a model can find a less common value from the text (Figure~\ref{fig:extended}, bottom). A model that leans towards frequent values scores low here regardless, so we compare each model with itself without a corpus (\textsc{name-only}). Adding a corpus lowers accuracy rather than raising it: at 16 pairs, Gemma~4 falls from 0.410 to 0.367 and Qwen3.6 from 0.466 to 0.355, and more pairs do not bring either model clearly above \textsc{name-only}.

\section{Discussion}
We now discuss implications for language documentation and research on metalinguistic abilities.

\subsection{Language Documentation}

Despite more than two decades of NLP work aimed at language documentation, documentary linguists rarely use it, partly because NLP projects assume an amount or type of data that linguists do not have~\citep{Gessler22-ClosingNLPGap, Gessler25-UnderstandingGapAnalysis}. Typological prediction is a case in point. It has mostly meant predicting features from other features of the same language~\citep{Bjerva20-SIGTYP2020Shared}, and in a survey by~\citet{bjerva-2024-role}, 34 linguists rate such predictions as not useful; what they ask for instead is prediction from text, with attributions to the examples in the data. TFI is set up for this case: its input is the kind of data documentation produces, and its answers can be checked against expert coding.

\citet{Gessler25-UnderstandingGapAnalysis} urge NLP researchers to state when a method would actually help linguists; our results suggest when an LLM would not. On languages it does not already know, the ones that need documenting, a model falls back on typological majority and on the pivot's grammar. LLM output on an unfamiliar language is thus best treated as a hypothesis for the linguist to check, especially when it matches the pivot's value.

This division of labour reflects a basic difference between the two. A field linguist starts from no knowledge of the language and checks hypotheses against the data; a model starts from broad knowledge of many languages and, as we find, uses it in place of the data whenever it can. Accuracy alone therefore cannot put the two on one scale.

\sa{I will also add somewhere here a few lines about how a) this means models differ from humans in metalinguistic abilities (being multilingual and having parametric knowledge to hand is hard to compare), and b) we shouldn't compare humans and llms anyway.}

\subsection{Metalinguistic Abilities of LLMs}
We posited that metalinguistic tasks are hard partly because they conflate inducing a rule with applying it. Our results show that induction alone is already hard: once recall is blocked, the models fall back on typological majority, much as models on constructed languages handle common patterns far better than rare ones~\citep{taguchi-sproat-2026-creating}. Part of the headroom on olympiad problems and grammar book translation~\citep{Aycock24-CanLLMsReallya, choudhary2025unveiling} may thus lie in finding the rule. The converse of TFI, which states the rule and asks the model to apply it~\citep{marmonier-etal-2025-explicit, yang-etal-2025-linggym}, would measure the other step, and paired with TFI it would locate a failure on either side of the rule.

Benchmarks built on natural text must also block recall, since models fall back on what they know of a language whenever they recognise it. Recent work therefore enciphers the data~\citep{bafna2026rashid, khouja2026lingolytoo}, but a cipher within the language's own alphabet leaves its diacritics as a fingerprint. Future benchmarks could remove recall more finely, for instance with models whose training data is open. The pivot also deserves closer study, since our results show that its grammar can pass for induction.

Tools may help with induction, much of which is mechanical work that a program does more reliably than a model in context: aligning words, retrieving sentences with a candidate marker and counting word orders. Typological signal can even be extracted from representations learned over parallel text~\citep{ostling-kurfali-2023-language}. Equipping models with such tools and asking them to cite the pairs behind each answer is a natural next step.

\section{Related Work}
\label{sec:lit}

\paragraph{Typology in NLP}
Typological databases such as WALS~\citep{wals} and Grambank~\citep{Skirgard23-GrambankV1} serve NLP as a source of features for cross-lingual transfer~\citep{ammar-etal-2016-many} and as a yardstick for the diversity of multilingual evaluation~\citep{Ploeger24-WhatTypologicalDiversity}. Completing typological databases is a long-standing task~\citep{ponti-etal-2019-modeling}, whether from other features~\citep{Bjerva20-SIGTYP2020Shared}, from parallel text through word alignment or learned language representations~\citep{ostling-2015-word, malaviya-etal-2017-learning}, from a language's name and its genealogical and areal neighbours~\citep{wang2026typological}, or from grammatical descriptions and glossed examples~\citep{hus-anastasopoulos-2026-rag, yang2026llm}. These approaches take the language's identity or a description of it as given, or process a whole corpus with a dedicated pipeline; TFI instead tests whether a model can induce a feature in context from raw text, as linguists have asked for~\citep{bjerva-2024-role}.

\paragraph{Metalinguistic Reasoning}
LLMs' ability to reason about language has been tested with linguistics olympiad puzzles~\citep{sahin-etal-2020-puzzling, bean2024lingoly, sanchez2026linguini, majmudar-filatova-2025-llms, choudhary2025unveiling}, gloss prediction from grammar snippets~\citep{yang-etal-2025-linggym}, and translation from a grammar book and dictionary~\citep{Tanzer24-BenchmarkLearningTranslatea, zhang-etal-2024-teaching, zhang-etal-2024-hire, hus-anastasopoulos-2024-back, zhang-etal-2025-read, Aycock24-CanLLMsReallya}. Related work asks whether models can learn from explicit grammatical explanations~\citep{marmonier-etal-2025-explicit} or benefit from step-by-step linguistic analyses~\citep{pei2026reasoning}. Since test languages may have been seen in pretraining, recent work constructs novel languages~\citep{taguchi-sproat-2026-creating}, or obfuscates natural languages with ciphers and permutations~\citep{bafna2026rashid, khouja2026lingolytoo}.
These all conflate inducing a rule with applying it; TFI isolates induction.

\section{Conclusion}
\label{sec:con}

We introduce the task of typological feature induction to isolate inductive metalinguistic reasoning. We find that models heavily rely on recall; on languages they cannot identify, they fail to beat a majority baseline across Grambank's features. This substantial headroom indicates models' inductive reasoning skills remain weak. We therefore emphasise to both linguists and NLP practitioners that such models are immature for direct application to real-world field linguistics tasks. We hope our findings stimulate further research into understanding, and improving LLMs' metalinguistic abilities.

\section*{Limitations}
\label{sec:lim}
Our corpora are FLORES+ and BOUQuET, chosen because they cover many languages with the same content. They consist of written text, most of it translated rather than originally composed in the target language, whereas documentation corpora are often transcribed speech or elicited sentences. Translation may bring the target text closer to the source language in places, but this cannot account for the pivot effect we report: the target text is identical across the 15 pivots, and answers follow whichever pivot is shown (\S\ref{sec:res-pivot}). Testing TFI on documentation corpora is a natural extension.

We treat a language as unknown when none of the models names it (\S\ref{sec:dis-extended}). A model may still recognise the family or region of such a language, so this set reduces rather than removes recall. Any remaining recall would raise accuracy, so it does not affect our finding that the models fail to beat the majority baseline.

Finally, some Grambank features may not be inducible from a given sample of sentences, as the corpora were not built to exhibit them. Our accuracies are therefore a lower bound on what a model could achieve with sufficient evidence. In future work, we plan to have linguists mark which features a sample of the data supports, which would allow precision and recall to be measured over inducible features only, closer to the task of a field linguist.

\bibliography{misc/references, misc/extra}
\bibliographystyle{misc/acl_natbib}

% \clearpage
\appendix

\section{Full Language List}
\label{app:lang-list}

Table~\ref{tab:langs-all} lists the 195 target languages with their ISO 639-3 code, Glottocode, language family, script (ISO 15924 codes) and source corpus, and marks which evaluation sets each belongs to. All 195 languages are in the core set; 127 are in the Latin-script set and 24 in the extended set. The languages come from 40 families, led by Indo-European (49), Atlantic-Congo (26), Austronesian (23) and Afro-Asiatic (14), and are written in 26 scripts.

\paragraph{Latin-script and extended sets.} 
The Latin-script set contains the Latin-script languages that the cipher (\S\ref{sec:met-controls}) can encipher. The cipher maps each letter that decomposes into a base letter and a diacritic onto a pool of precomposed Latin letters, and permutes the letters without a base letter (seven letters such as \textit{\ae}, \textipa{N} and \textit{\oe}) among themselves. 
Seven of the 134 Latin-script languages need more letters than this pool holds and are excluded. Ewe, Fon, Kxoe, Hassaniyya and Dar Daju Daju use many letters without a base letter, such as \textipa{E}, \textipa{O} and click letters, and Vietnamese and Duala add tone marks to vowels that already carry other diacritics.
The extended set contains the 24 languages that none of the three models identifies (\S\ref{sec:met-eval}): Arhuaco, Bafia, Brahui, Burushaski, Central Kanuri, Dandami Maria, Dar Daju Daju, Duala, Dyula, Eton (Cameroon), Fur, Hassaniyya, Hehe, Koyra Chiini Songhay, Kxoe, Lendu, Lugbara, Mai Brat, Ngambay, Nuer, Siamou, Western Cham, Xârâcùù and Yemsa. Fifteen of them are in the Latin-script set.

\paragraph{Matching corpus varieties to Grambank.} Grambank does not list ISO 639-3 codes, so corpus languages are matched to Grambank entries through Glottolog. Grambank codes a few languages at the level of a macrolanguage while the corpora use the codes of its members; we assign the macrolanguage's values to the member that is its standard variety, giving Bosnian, Croatian and Serbian the values of Serbo-Croatian (\textit{sout1528}) and Standard Latvian those of Latvian (\textit{latv1249}). Where FLORES+ has several dialect versions of one language, we keep only the variety that Grambank codes (North Levantine Arabic, standard Catalan). Twi, for which Grambank codes only the Akan macrolanguage, is excluded. Each target uses a single script.

\begin{table}[h]
\centering\footnotesize
\setlength\tabcolsep{3pt}
\begin{tabular}{lrrrl}
\toprule
 & Lang. & Feat. & Items & Used for \\
\midrule
Core & 195 & 22 & 4,084 & \S\ref{sec:res-saturate} \\
Latin-Script & 127 & 22 & 2,657 & \S\ref{sec:res-langid}--\S\ref{sec:res-pivot} \\
Extended & 24 & 195 & 3,775 & \S\ref{sec:dis-extended} \\
\bottomrule
\end{tabular}
\caption{Statistics of the three evaluation sets. Feat.\ is the number of distinct Grambank features asked about. The extended set asks about every feature coded for each of its languages.}
\label{tab:benchmarks}
\end{table}

\section{Features}
\label{app:features}

\paragraph{Statistics} The statistics of three evaluation sets are shown in Table~\ref{tab:benchmarks}.

\paragraph{List of Core Features}
Table~\ref{tab:core-features} lists the 22 Grambank features in the core set, grouped into word order, word classes and morphology following Grambank's own feature groups. We chose features whose value shows up in word order or overt marking within a sentence, so that a sentence-level corpus can in principle decide them. Not every feature is coded for every language, so the number of items per feature ranges from 162 to 194. Some features are strongly skewed: 180 of 186 languages lack demonstrative classifiers (GB038), and 157 of 194 have SV order in intransitive clauses (GB130).

\section{Implementation Details}
\label{app:impl}

\subsection{Prompt Templates}
\label{app:prompts}

\newcommand{\promptbox}[3]{%
  \par\noindent\colorbox{#2!15}{\textbf{\scriptsize #1}}\par\vspace{-2pt}%
  \noindent\fcolorbox{#2!45!black}{#2!4}{\parbox{0.96\linewidth}{\vspace{2pt}\small #3\par\vspace{2pt}}}\par\vspace{6pt}%
}
\newcommand{\slot}[1]{\textit{\{#1\}}}

Each item is sent as a single user turn through the model's chat template, with no system prompt. The corpus comes first and the question second, so the 22 items of a language share the same prompt prefix. Answer options use Grambank's numeric value codes and are listed in a fixed random order per item; the answer is read from the final \texttt{Answer:} line.

\paragraph{Question} Every setting ends with the same question block. \slot{question} is Grambank's question and \slot{summary} the Summary section of its feature description. We do not use the Examples section, which names languages together with their values.

\promptbox{Question block}{teal}{%
You are a linguist answering a question about the grammar of a specific language.\\[2pt]
[Language: \slot{language} (ISO 639-3: \slot{iso}) \quad \textit{only in \textsc{name-only} and \textsc{+name}}]\\[2pt]
Question: \slot{question}\\[2pt]
What this question means:\\
\slot{summary}\\[2pt]
Options:\\
\hspace*{1em}\slot{code} = \slot{label}\\
\hspace*{1em}\dots\\[2pt]
\slot{instruction} You must choose exactly one option.\\
End your reply with a final line in the form: Answer: <\slot{codes}>%
}

Here is an example of Polish for the feature GB024, with the corpus shortened to one pair:

\promptbox{\textsc{parallel}, Polish, GB024}{black}{%
Below is a parallel corpus between English and an unnamed target language. Each pair is one sentence: EN is English, TL is the target language.\\[2pt]
EN: 28-year-old Vidal had joined Barça three seasons ago, from Sevilla.\\
TL: 28-letni Vidal przyszedł do Barcelony z Sewilli trzy sezony temu.\\
{[}\dots\ $n$ pairs in total{]}\\[2pt]
-{}-{}-\\[2pt]
You are a linguist answering a question about the grammar of a specific language.\\[2pt]
Question: What is the order of numeral and noun in the NP?\\[2pt]
What this question means:\\
This feature focuses on the order of cardinal numerals (not ordinal numerals) and nouns in quantified noun phrases. In some languages only one order (Numeral-Noun or Noun-Numeral) is possible. In other languages both orders may be possible. This can take the form of free variation (where Numeral-Noun and Noun-Numeral order can occur interchangeably). The order may also depend on some other grammatical category which is neutral with respect to pragmaticality (e.g. definiteness). If the order depends on the noun, consider only the orders that apply to an open class of nouns and disregard orders that are restricted to a specific closed class of nouns.\\[2pt]
Options:\\
\hspace*{1em}2 = N-Num\\
\hspace*{1em}1 = Num-N\\
\hspace*{1em}3 = both\\[2pt]
Answer based only on the corpus above. You must choose exactly one option.\\
End your reply with a final line in the form: Answer: <2/1/3>%
}

GPT-5.6 Luna reasons before answering, but the API does not return this reasoning, so its outputs would otherwise contain little more than the answer line. To see what evidence it uses, we ask it to write out its analysis. In the plaintext settings and \textsc{name-only}, the sentence \textit{End your reply with a final line in the form:} is replaced by \textit{First write your step-by-step analysis of the evidence, quoting the specific sentences you rely on. Then end with a final line in the form:} (in \textsc{name-only}: \textit{First write your step-by-step reasoning. Then end with a final line in the form:}).

\paragraph{Language Identification} The probe uses the header and corpus of \textsc{parallel} (or \textsc{cipher-parallel}), followed by:

\promptbox{Language identification}{teal}{%
-{}-{}-\\[2pt]
What is the target language (TL)? Reply with your single best guess.\\
If you are unsure, still give your best guess.\\
End your reply with a final line in the form: Language: <name>%
}

\subsection{Hyperparameters}
Open models run on vLLM~\citep{kwon2023efficient} in bfloat16 with each model's recommended sampling parameters (temperature 1.0, top-$p$ 0.95; top-$k$ 64 for Gemma, 20 for Qwen), thinking mode off, and a 12,288-token output budget.
Open models ran over four NVIDIA A100 40GB GPUs per job. All experiments, including pilot runs, took about 2,000 GPU hours.
GPT-5.6 Luna uses server defaults with a 4,096-token reasoning budget; in the plaintext settings its prompt additionally asks it to explain its answer.

Our open models run with thinking mode off, although they still reason step by step in their outputs. To check that this does not understate their ability, we ran both models with thinking on in \textsc{cipher-parallel} at $n{=}128$ on the first corpus seed, with a 32,768-token output budget and otherwise the same settings. Thinking makes outputs 5--14 times longer but does not raise overall accuracy for either model: 0.622 with thinking vs.\ 0.636 without for Qwen3.6, and 0.644 in both cases for Gemma~4, and the two runs give the same answer on 83\% and 79\% of items.

\subsection{Artifacts and Licenses}
Grambank and Glottolog are released under CC BY 4.0, FLORES+ and BOUQuET under CC BY-SA 4.0. FLORES+ and BOUQuET require accepting their terms of use before download, so we release only item definitions, corpus sentence IDs and code, not the text. Gemma~4, Qwen3.6 and gpt-oss-20b are used under their published model licenses, and GPT-5.6 Luna through its API. All are used for research evaluation, in line with their intended use.

\section{Unanswered items}
\label{app:unparsed}

\begin{table*}[t]
\centering\small
\setlength\tabcolsep{4pt}
\begin{tabular}{llccccccccc}
\toprule
 & & \multicolumn{3}{c}{$n{=}128$} & \multicolumn{3}{c}{$n{=}512$} & \multicolumn{3}{c}{$n{=}1024$} \\
\cmidrule(lr){3-5}\cmidrule(lr){6-8}\cmidrule(lr){9-11}
 & & all & answered & unans. & all & answered & unans. & all & answered & unans. \\
\midrule
\multirow{3}{*}{Qwen3.6-27B} & \textsc{parallel} & 0.788 & 0.789 & 5 & 0.777 & 0.786 & 47 & 0.775 & 0.784 & 51 \\
 & \textsc{mono} & 0.790 & 0.792 & 11 & 0.749 & 0.777 & 146 & 0.734 & 0.770 & 191 \\
 & \textsc{misaligned} & 0.787 & 0.792 & 29 & 0.749 & 0.786 & 193 & 0.746 & 0.784 & 198 \\
\midrule
\multirow{3}{*}{Gemma~4~31B} & \textsc{parallel} & 0.807 & 0.807 & 0 & 0.810 & 0.810 & 1 & 0.810 & 0.812 & 7 \\
 & \textsc{mono} & 0.803 & 0.803 & 1 & 0.804 & 0.805 & 3 & 0.800 & 0.801 & 8 \\
 & \textsc{misaligned} & 0.801 & 0.801 & 0 & 0.799 & 0.799 & 2 & 0.792 & 0.795 & 16 \\
\bottomrule
\end{tabular}
\caption{Accuracy on all 4,084 items (unanswered counted as wrong, as in the main text), accuracy on answered items only, and the number of unanswered items per run. Mean over three corpus seeds.}
\label{tab:unparsed}
\end{table*}

All accuracies in the main text count an unanswered item as wrong. Table~\ref{tab:unparsed} separates the two components. For Gemma~4 the distinction is immaterial: at most 16 of the 4,084 items per run go unanswered, and accuracy on answered items is within 0.3 points of accuracy on all items. For Qwen3.6 it is not. Between 128 and 1,024 pairs, the number of unanswered items per run rises from 5 to 51 under \textsc{parallel} and from 11--29 to 191--198 under \textsc{mono} and \textsc{misaligned}. Accuracy on answered items changes much less: from 0.789 to 0.784 under \textsc{parallel}, from 0.792 to 0.784 under \textsc{misaligned}, and from 0.792 to 0.770 under \textsc{mono}. The decline of Qwen3.6 with corpus size is therefore mostly a decline in the rate at which it answers.

\paragraph{Runaway Outputs}
Almost all unanswered items are outputs that reach the 12,288-token limit. They are rarely verbatim loops; typically the model works through corpus sentences one by one (``Let's look at: EN: \dots\ TL: \dots'') without reaching a conclusion. They concentrate on features that require locating the verb or the negator in every sentence (GB130--GB133, GB137, GB138), and more so without aligned translations: at 512 pairs, 3.5\% of these items under \textsc{parallel}, 9.7\% under \textsc{mono} and 12.7\% under \textsc{misaligned}. Since the outputs show no sign of converging, a larger budget is unlikely to recover them, and we count them as wrong.
\paragraph{Context Overflow}
At 1,024 pairs the Santali corpus (Ol Chiki script) exceeds Qwen3.6's 262,144-token context window, so its 21 items return no output in every run (0.5 points). This is the only case of context overflow in the experiments.

\section{Language Naming in Reasoning}
\label{app:langname}

We use gpt-oss-20b with low reasoning effort and greedy decoding, so that the judge is neither of the evaluated models. For each output it extracts the most specific language, family or region that the output says the target is, together with a supporting quote. Languages mentioned for comparison (``unlike English''), the pivot language, and names that occur inside quoted corpus text do not count. Whether an extracted name is the target is decided by the same matcher as Table~\ref{tab:langid}. We judged the outputs of the first corpus seed at $n \in \{16, 128\}$, 21,256 outputs in total.

In a pilot on 100 outputs, the judge sometimes recognised the language itself from quoted corpus words although the output named none, so we discard any extracted name that does not occur in the output. A random sample of 40 remaining extracted names were all correct.

Table~\ref{tab:langname-split} splits accuracy by whether the plaintext output names the target correctly.

\begin{table}[h]
\centering\small
\setlength\tabcolsep{4pt}
\begin{tabular}{llcc}
\toprule
 & & Named correctly & Not named \\
\midrule
\multirow{2}{*}{Gemma~4} & $n{=}16$ & .784 $\rightarrow$ .620 & .760 $\rightarrow$ .612 \\
 & $n{=}128$ & .821 $\rightarrow$ .676 & .761 $\rightarrow$ .623 \\
\midrule
\multirow{2}{*}{Qwen3.6} & $n{=}16$ & .793 $\rightarrow$ .588 & .715 $\rightarrow$ .642 \\
 & $n{=}128$ & .810 $\rightarrow$ .646 & .724 $\rightarrow$ .638 \\
\bottomrule
\end{tabular}
\caption{Accuracy in \textsc{parallel} $\rightarrow$ \textsc{cipher-parallel} on the Latin-script items, split by whether the plaintext output names the target correctly (first corpus seed).}
\label{tab:langname-split}
\end{table}

\section{Further Analyses of Pivot Swap}
\label{app:pivot}

\paragraph{Exceptions in Figure~\ref{fig:pivot}.}
English and French with Gemma~4 go the same way as the other pivots but miss significance on the disagreeing items; only Gemma~4 with Mandarin and Qwen3.6 with Hungarian show no effect on either side.

\paragraph{Choice model.}
Each option of each answer is one row (three answers per item, one per corpus seed). A linear probability model with a fixed effect per answer predicts whether the option was chosen from three indicators: it is the target's value, the pivot's own value, or the majority value across Grambank. We fit it on \textsc{cipher-mono} and \textsc{cipher-parallel} together, letting each coefficient change in \textsc{cipher-parallel}, with standard errors clustered by target and pivot. \textsc{cipher-mono} never shows the pivot, so its pivot coefficient (10 points for Gemma~4, 11 for Qwen3.6) only measures how often the pivot's value is what the models say anyway, and the effect of the pivot side is the change from \textsc{cipher-mono} to \textsc{cipher-parallel}. The pivot side raises the coefficient of the pivot's value by 20.0 points for Gemma~4 ($t{=}7.9$) and 14.4 for Qwen3.6 ($t{=}8.0$), and changes that of the target's value by $-$0.4 ($t{=}{-}0.3$) and $-$2.3 ($t{=}{-}1.8$). A conditional logit gives the same result (change in the pivot coefficient $+$0.49 and $+$0.37 log-odds, in the target coefficient $+$0.04 and $-$0.01). With a fixed effect for each target and feature, a matching pivot raises accuracy as much when its value is not the majority one.

\begin{figure}[t]
\centering
\includegraphics{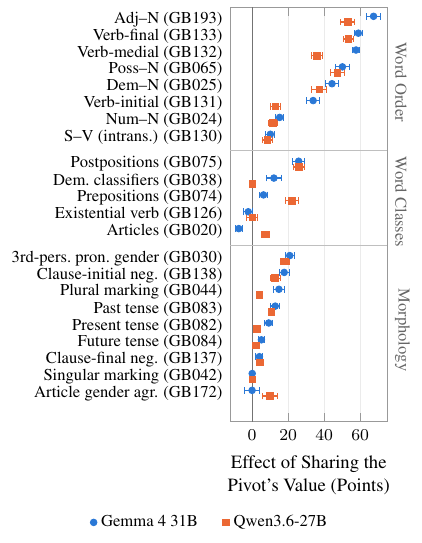}
\caption{Change in accuracy when the pivot shares the target's value for the feature, with a fixed effect for each target and feature, so that pivots are compared on the same question ($n{=}128$, 15 pivots). Bars are 95\% confidence intervals with standard errors clustered by target and pivot.}
\label{fig:pivot-feature}
\end{figure}

\paragraph{Word order.}
Comparing pivots on the same target and feature, a pivot that shares the target's value raises accuracy by 46 points on word-order features for Gemma~4 and 36 for Qwen3.6, against 6--14 on word classes and 7--10 on morphology (Figure~\ref{fig:pivot-feature}). The effect is largest for adjective--noun order and for verb-final and verb-medial order, up to 67 points, and close to zero for singular marking and the existential verb. Where the target differs from the pivot, the pivot's value is given in 44\% of Gemma~4's word-order answers without a pivot side and 65\% with it, 38\% of the way from no pivot effect to always copying; on morphology the share barely moves (43\% to 47\%).

\paragraph{Overall effect of each pivot.}
Because each pivot helps where the target shares its value and hurts where it does not, its overall effect depends on how often the targets happen to share its values. Verb-medial pivots, whose word order matches most targets, give overall gains of up to 4.8 points (English, Gemma~4), while Turkish, Korean, Hindi and Japanese give losses of 1.8--4.6 points with both models. English owes its advantage to agreement: 62\% of items share English's value, against 50\% for Japanese.

\paragraph{What explains the pivot's effect.}
Fixed effects for each target and feature explain 63\% of the variance in item accuracy for Gemma~4 and 73\% for Qwen3.6, against a ceiling of 83\% and 86\% set by sampling noise over the three corpus seeds. Of the remaining variance that depends on the pivot and is not noise, the identity of the pivot explains 2\%, the syntactic distance between pivot and target (lang2vec~\cite{littell-etal-2017-uriel}) 8\%, whether the pivot shares the target's value 27--29\%, and that agreement with a separate effect per feature 54--57\%. Once syntactic distance is controlled, the verb-final penalty above shrinks to 1.1--1.5 points and is no longer significant. General similarity adds nothing beyond agreement on the tested features: regressed on the 1,899 (target, pivot) pairs, agreement on the remaining Grambank features predicts the gain on its own (Gemma~4: $\beta{=}0.29$, $t{=}8.3$; Qwen3.6: $\beta{=}0.18$, $t{=}5.5$), but not once agreement on the tested features is in the model ($\beta{=}{-}0.01$, $t{=}{-}0.3$; $\beta{=}{-}0.07$, $t{=}{-}2.2$), which remains strong ($t{=}19.9$ and $17.4$).

\paragraph{Choosing the pivot.}
Choosing the syntactically closest pivot for each target would raise accuracy from 64.2\% to 68.7\% for Gemma~4 and from 62.1\% to 67.2\% for Qwen3.6 over English, as much as the best pivot chosen on held-out seeds. Closeness is measured on the target's documented grammar, which is what the task asks for, so this is not a usable strategy but another sign that the gain comes from agreement.

\begin{table}[h]
\centering\small
\setlength\tabcolsep{4pt}
\begin{tabular}{lcccc}
\toprule
 & \multicolumn{2}{c}{Gemma~4} & \multicolumn{2}{c}{Qwen3.6} \\
\cmidrule(lr){2-3}\cmidrule(lr){4-5}
 & Unid. & Others & Unid. & Others \\
\midrule
\textsc{majority} & .676 & .603 & .676 & .603 \\
\textsc{parallel} & .695 & .798 & .701 & .775 \\
\textsc{cipher-parallel} & .614 & .646 & .626 & .630 \\
\bottomrule
\end{tabular}
\caption{Accuracy on the Latin-script core items at $n{=}128$, mean over three corpus seeds, split by whether the language is unidentified, i.e.\ belongs to the extended set (15 languages, 318 items), or not (112 languages, 2,339 items).}
\label{tab:cipher-cost}
\end{table}

\section{Cost of the Cipher}
\label{app:cipher-cost}

The cipher is meant to remove recall, but it also changes how the text is tokenized and removes any cue carried by word forms. To separate the two, Table~\ref{tab:cipher-cost} splits the Latin-script core items by whether the language is one that none of the models can identify (the 15 Latin-script languages of the extended set). On these languages there is little to recall, yet enciphering lowers accuracy by 8.1 points for Gemma~4 and 7.5 for Qwen3.6, against 15.2 and 14.5 on the other languages. Roughly half of the drop under the cipher therefore does not come from lost recall. Plaintext word forms may still reveal a family or region, or enciphered text may simply be harder to process; our data do not separate the two. Either way, the cost is shared by all enciphered settings, so the comparisons in \S\ref{sec:res-cipher} and \S\ref{sec:res-pivot}, which are all between enciphered settings, are not affected. On these 15 languages, plaintext accuracy is only 2 points above \textsc{majority}, in line with the extended set (\S\ref{sec:dis-extended}).

\begin{table*}[t]
\centering\scriptsize
\begin{tabular}{@{}l@{\hspace{6pt}}l@{\hspace{6pt}}l@{\hspace{6pt}}l@{\hspace{6pt}}l@{\hspace{6pt}}l@{\hspace{6pt}}r@{\hspace{6pt}}c@{\hspace{4pt}}c@{}}
\toprule
Language & ISO & Glottocode & Family & Script & Src. & Core & L & E \\
\midrule
Achinese & ace & achi1257 & Austronesian & Latn & F & 22 & \checkmark &  \\
Afar & aar & afar1241 & Afro-Asiatic & Latn & B & 21 & \checkmark &  \\
Aguaruna & agr & agua1253 & Chicham & Latn & B & 20 & \checkmark &  \\
Amharic & amh & amha1245 & Afro-Asiatic & Ethi & F+B & 21 &  &  \\
Amis & ami & amis1246 & Austronesian & Latn & B & 22 & \checkmark &  \\
Arhuaco & arh & arhu1242 & Chibchan & Latn & B & 22 & \checkmark & \checkmark \\
Armenian & hye & nucl1235 & Indo-European & Armn & F+B & 21 &  &  \\
Awadhi & awa & awad1243 & Indo-European & Deva & F & 20 &  &  \\
Ayacucho Quechua & quy & ayac1239 & Quechuan & Latn & F & 21 & \checkmark &  \\
Bafia & ksf & bafi1243 & Atlantic-Congo & Latn & B & 20 & \checkmark & \checkmark \\
Balinese & ban & bali1278 & Austronesian & Latn & F & 22 & \checkmark &  \\
Bambara & bam & bamb1269 & Mande & Latn & F+B & 18 & \checkmark &  \\
Bashkir & bak & bash1264 & Turkic & Cyrl & F+B & 21 &  &  \\
Basque & eus & basq1248 & Isolate & Latn & F+B & 22 & \checkmark &  \\
Belarusian & bel & bela1254 & Indo-European & Cyrl & F+B & 22 &  &  \\
Bemba (Zambia) & bem & bemb1257 & Atlantic-Congo & Latn & F & 22 & \checkmark &  \\
Bhojpuri & bho & bhoj1244 & Indo-European & Deva & F+B & 21 &  &  \\
Bodo (India) & brx & bodo1269 & Sino-Tibetan & Deva & F+B & 20 &  &  \\
Borana-Arsi-Guji Oromo & gax & bora1271 & Afro-Asiatic & Latn & B & 18 & \checkmark &  \\
Bosnian & bos & sout1528 & Indo-European & Latn & F+B & 22 & \checkmark &  \\
Brahui & brh & brah1256 & Dravidian & Arab & B & 22 &  & \checkmark \\
Breton & bre & bret1244 & Indo-European & Latn & B & 22 & \checkmark &  \\
Buginese & bug & bugi1244 & Austronesian & Latn & F & 22 & \checkmark &  \\
Burmese & mya & nucl1310 & Sino-Tibetan & Mymr & F+B & 22 &  &  \\
Burushaski & bsk & buru1296 & Isolate & Arab & B & 22 &  & \checkmark \\
Campidanese Sardinian & sro & camp1261 & Indo-European & Latn & B & 21 & \checkmark &  \\
Catalan & cat & stan1289 & Indo-European & Latn & F+B & 22 & \checkmark &  \\
Cebuano & ceb & cebu1242 & Austronesian & Latn & F+B & 18 & \checkmark &  \\
Central Aymara & ayr & cent2142 & Aymaran & Latn & F+B & 22 & \checkmark &  \\
Central Kanuri & knc & cent2050 & Saharan & Arab & F+B & 22 &  & \checkmark \\
Central Kurdish & ckb & cent1972 & Indo-European & Arab & F+B & 1 &  &  \\
Chechen & che & chec1245 & Nakh-Daghestanian & Cyrl & B & 22 &  &  \\
Cherokee & chr & cher1273 & Iroquoian & Cher & B & 22 &  &  \\
Chuvash & chv & chuv1255 & Turkic & Cyrl & F+B & 22 &  &  \\
Crimean Tatar & crh & crim1257 & Turkic & Latn & F & 22 & \checkmark &  \\
Croatian & hrv & sout1528 & Indo-European & Latn & F+B & 22 & \checkmark &  \\
Cusco Quechua & quz & cusc1236 & Quechuan & Latn & B & 22 & \checkmark &  \\
Czech & ces & czec1258 & Indo-European & Latn & F+B & 22 & \checkmark &  \\
Dandami Maria & daq & dand1238 & Dravidian & Deva & B & 22 &  & \checkmark \\
Danish & dan & dani1285 & Indo-European & Latn & F+B & 22 & \checkmark &  \\
Dar Daju Daju & djc & dard1243 & Dajuic & Latn & B & 12 &  & \checkmark \\
Dargwa & dar & darg1241 & Nakh-Daghestanian & Cyrl & F & 19 &  &  \\
Duala & dua & dual1243 & Atlantic-Congo & Latn & B & 20 &  & \checkmark \\
Dutch & nld & dutc1256 & Indo-European & Latn & F+B & 22 & \checkmark &  \\
Dyula & dyu & dyul1238 & Mande & Latn & F & 20 & \checkmark & \checkmark \\
Dzongkha & dzo & dzon1239 & Sino-Tibetan & Tibt & F+B & 22 &  &  \\
Eastern Mari & mhr & gras1239 & Uralic & Cyrl & F & 22 &  &  \\
Egyptian Arabic & arz & egyp1253 & Afro-Asiatic & Arab & F+B & 22 &  &  \\
Erzya & myv & erzy1239 & Uralic & Cyrl & F & 22 &  &  \\
Eton (Cameroon) & eto & eton1253 & Atlantic-Congo & Latn & B & 21 & \checkmark & \checkmark \\
Ewe & ewe & ewee1241 & Atlantic-Congo & Latn & F & 20 &  &  \\
Faroese & fao & faro1244 & Indo-European & Latn & F+B & 22 & \checkmark &  \\
Fijian & fij & fiji1243 & Austronesian & Latn & F & 22 & \checkmark &  \\
Filipino & fil & fili1244 & Austronesian & Latn & F & 22 & \checkmark &  \\
Finnish & fin & finn1318 & Uralic & Latn & F+B & 22 & \checkmark &  \\
Fon & fon & fonn1241 & Atlantic-Congo & Latn & F & 22 &  &  \\
French & fra & stan1290 & Indo-European & Latn & F+B & 22 & \checkmark &  \\
Fur & fvr & furr1244 & Furan & Latn & B & 22 & \checkmark & \checkmark \\
Galician & glg & gali1258 & Indo-European & Latn & F+B & 22 & \checkmark &  \\
Ganda & lug & gand1255 & Atlantic-Congo & Latn & F+B & 22 & \checkmark &  \\
Georgian & kat & nucl1302 & Kartvelian & Geor & F+B & 22 &  &  \\
Gilbertese & gil & gilb1244 & Austronesian & Latn & B & 22 & \checkmark &  \\
Gusii & guz & gusi1247 & Atlantic-Congo & Latn & B & 21 & \checkmark &  \\
Haitian & hat & hait1244 & Indo-European & Latn & F+B & 21 & \checkmark &  \\
Halh Mongolian & khk & halh1238 & Mongolic-Khitan & Cyrl & F+B & 22 &  &  \\
\bottomrule
\end{tabular}
\caption{All 195 target languages, with family from Grambank. Src.\ is the source corpus (F: FLORES+, B: BOUQuET); Core is the number of core-set items; L and E mark the Latin-script and extended sets.}
\label{tab:langs-all}
\end{table*}
\begin{table*}[p]
\centering\scriptsize
\begin{tabular}{@{}l@{\hspace{6pt}}l@{\hspace{6pt}}l@{\hspace{6pt}}l@{\hspace{6pt}}l@{\hspace{6pt}}l@{\hspace{6pt}}r@{\hspace{6pt}}c@{\hspace{4pt}}c@{}}
\toprule
Language & ISO & Glottocode & Family & Script & Src. & Core & L & E \\
\midrule
Hassaniyya & mey & hass1238 & Afro-Asiatic & Latn & B & 22 &  & \checkmark \\
Hausa & hau & haus1257 & Afro-Asiatic & Latn & F+B & 19 & \checkmark &  \\
Hebrew & heb & hebr1245 & Afro-Asiatic & Hebr & F+B & 22 &  &  \\
Hehe & heh & hehe1240 & Atlantic-Congo & Latn & B & 21 & \checkmark & \checkmark \\
Highland Puebla Nahuatl & azz & high1278 & Uto-Aztecan & Latn & B & 14 & \checkmark &  \\
Hindi & hin & hind1269 & Indo-European & Deva & F+B & 22 &  &  \\
Hungarian & hun & hung1274 & Uralic & Latn & F+B & 22 & \checkmark &  \\
Icelandic & isl & icel1247 & Indo-European & Latn & F+B & 22 & \checkmark &  \\
Igbo & ibo & nucl1417 & Atlantic-Congo & Latn & F+B & 22 & \checkmark &  \\
Iloko & ilo & ilok1237 & Austronesian & Latn & F+B & 20 & \checkmark &  \\
Indonesian & ind & indo1316 & Austronesian & Latn & F+B & 19 & \checkmark &  \\
Iranian Persian & pes & west2369 & Indo-European & Arab & F+B & 22 &  &  \\
Iraqw & irk & iraq1241 & Afro-Asiatic & Latn & B & 21 & \checkmark &  \\
Irish & gle & iris1253 & Indo-European & Latn & F+B & 22 & \checkmark &  \\
Italian & ita & ital1282 & Indo-European & Latn & F+B & 22 & \checkmark &  \\
Japanese & jpn & nucl1643 & Japonic & Jpan & F+B & 22 &  &  \\
Javanese & jav & java1254 & Austronesian & Latn & F+B & 20 & \checkmark &  \\
K'iche' & quc & kich1262 & Mayan & Latn & B & 22 & \checkmark &  \\
Kabyle & kab & kaby1243 & Afro-Asiatic & Latn & F & 10 & \checkmark &  \\
Kachin & kac & kach1280 & Sino-Tibetan & Latn & F+B & 22 & \checkmark &  \\
Kalaallisut & kal & kala1399 & Eskimo-Aleut & Latn & B & 22 & \checkmark &  \\
Kamba (Kenya) & kam & kamb1297 & Atlantic-Congo & Latn & F+B & 21 & \checkmark &  \\
Kannada & kan & nucl1305 & Dravidian & Knda & F+B & 22 &  &  \\
Kaqchikel & cak & kaqc1270 & Mayan & Latn & B & 22 & \checkmark &  \\
Kashmiri & kas & kash1277 & Indo-European & Arab & F & 22 &  &  \\
Khakas & kjh & khak1248 & Turkic & Cyrl & F & 21 &  &  \\
Khmer & khm & cent1989 & Austroasiatic & Khmr & F+B & 22 &  &  \\
Khoekhoe & naq & nama1264 & Khoe-Kwadi & Latn & B & 18 & \checkmark &  \\
Kikuyu & kik & kiku1240 & Atlantic-Congo & Latn & F & 22 & \checkmark &  \\
Kipsigis & sgc & kips1239 & Nilotic & Latn & B & 22 & \checkmark &  \\
Korean & kor & kore1280 & Koreanic & Kore & F+B & 22 &  &  \\
Koyra Chiini Songhay & khq & koyr1240 & Songhay & Latn & B & 22 & \checkmark & \checkmark \\
Kurukh & kru & kuru1302 & Dravidian & Deva & B & 18 &  &  \\
Kxoe & xuu & kxoe1243 & Khoe-Kwadi & Latn & B & 22 &  & \checkmark \\
Lao & lao & laoo1244 & Tai-Kadai & Laoo & F+B & 19 &  &  \\
Lendu & led & lend1245 & Central Sudanic & Latn & B & 22 & \checkmark & \checkmark \\
Levantine Arabic & apc & nort3139 & Afro-Asiatic & Arab & F+B & 22 &  &  \\
Lingala & lin & ling1263 & Atlantic-Congo & Latn & F+B & 17 & \checkmark &  \\
Lithuanian & lit & lith1251 & Indo-European & Latn & F+B & 22 & \checkmark &  \\
Lombard & lmo & lomb1257 & Indo-European & Latn & F & 22 & \checkmark &  \\
Lugbara & lgg & lugb1240 & Central Sudanic & Latn & B & 22 & \checkmark & \checkmark \\
Luo (Kenya and Tanzania) & luo & luok1236 & Nilotic & Latn & F+B & 22 & \checkmark &  \\
Lushai & lus & lush1249 & Sino-Tibetan & Latn & F & 22 & \checkmark &  \\
Macedonian & mkd & mace1250 & Indo-European & Cyrl & F+B & 22 &  &  \\
Magahi & mag & maga1260 & Indo-European & Deva & F & 20 &  &  \\
Mai Brat & ayz & maib1239 & Isolate & Latn & B & 22 & \checkmark & \checkmark \\
Maithili & mai & mait1250 & Indo-European & Deva & F+B & 22 &  &  \\
Makhuwa & vmw & makh1264 & Atlantic-Congo & Latn & F+B & 21 & \checkmark &  \\
Malayalam & mal & mala1464 & Dravidian & Mlym & F+B & 22 &  &  \\
Maltese & mlt & malt1254 & Afro-Asiatic & Latn & F+B & 22 & \checkmark &  \\
Mam & mam & mamm1241 & Mayan & Latn & B & 22 & \checkmark &  \\
Mandarin Chinese & cmn & mand1415 & Sino-Tibetan & Hans & F+B & 22 &  &  \\
Maori & mri & maor1246 & Austronesian & Latn & F+B & 22 & \checkmark &  \\
Mapudungun & arn & mapu1245 & Araucanian & Latn & B & 22 & \checkmark &  \\
Marathi & mar & mara1378 & Indo-European & Deva & F+B & 22 &  &  \\
Masai & mas & masa1300 & Nilotic & Latn & B & 18 & \checkmark &  \\
Minangkabau & min & mina1268 & Austronesian & Arab & F+B & 22 &  &  \\
Modern Greek (1453-) & ell & mode1248 & Indo-European & Grek & F+B & 22 &  &  \\
Mískito & miq & misk1235 & Misumalpan & Latn & B & 21 & \checkmark &  \\
Nepali & npi & nepa1254 & Indo-European & Deva & F+B & 22 &  &  \\
Ngambay & sba & ngam1268 & Central Sudanic & Latn & B & 22 & \checkmark & \checkmark \\
Nigerian Fulfulde & fuv & nige1253 & Atlantic-Congo & Latn & F+B & 21 & \checkmark &  \\
North Azerbaijani & azj & nort2697 & Turkic & Latn & F+B & 22 & \checkmark &  \\
Northern Sami & sme & nort2671 & Uralic & Latn & B & 20 & \checkmark &  \\
Northern Uzbek & uzn & nort2690 & Turkic & Latn & F+B & 22 & \checkmark &  \\
\bottomrule
\end{tabular}
\caption{All 195 target languages (continued).}
\end{table*}
\begin{table*}[p]
\centering\scriptsize
\begin{tabular}{@{}l@{\hspace{6pt}}l@{\hspace{6pt}}l@{\hspace{6pt}}l@{\hspace{6pt}}l@{\hspace{6pt}}l@{\hspace{6pt}}r@{\hspace{6pt}}c@{\hspace{4pt}}c@{}}
\toprule
Language & ISO & Glottocode & Family & Script & Src. & Core & L & E \\
\midrule
Nuer & nus & nuer1246 & Nilotic & Latn & F+B & 20 & \checkmark & \checkmark \\
Occitan (post 1500) & oci & occi1239 & Indo-European & Latn & F & 22 & \checkmark &  \\
Odia & ory & oriy1255 & Indo-European & Orya & F+B & 22 &  &  \\
Pangasinan & pag & pang1290 & Austronesian & Latn & F & 22 & \checkmark &  \\
Panjabi & pan & panj1256 & Indo-European & Guru & F+B & 22 &  &  \\
Paraguayan Guaraní & gug & para1311 & Tupian & Latn & F+B & 10 & \checkmark &  \\
Plains Cree & crk & plai1258 & Algic & Latn & B & 19 & \checkmark &  \\
Plateau Malagasy & plt & plat1254 & Austronesian & Latn & F+B & 19 & \checkmark &  \\
Polish & pol & poli1260 & Indo-European & Latn & F+B & 22 & \checkmark &  \\
Portuguese & por & port1283 & Indo-European & Latn & F & 22 & \checkmark &  \\
Pulaar & fuc & pula1263 & Atlantic-Congo & Latn & B & 19 & \checkmark &  \\
Purepecha & tsz & pure1242 & Tarascan & Latn & B & 22 & \checkmark &  \\
Romansh & roh & roma1326 & Indo-European & Latn & B & 21 & \checkmark &  \\
Russian & rus & russ1263 & Indo-European & Cyrl & F+B & 22 &  &  \\
Samoan & smo & samo1305 & Austronesian & Latn & F & 22 & \checkmark &  \\
San Dionisio Del Mar Huave & hve & sand1278 & Huavean & Latn & B & 22 & \checkmark &  \\
Santali & sat & sant1410 & Austroasiatic & Olck & F+B & 21 &  &  \\
Serbian & srp & sout1528 & Indo-European & Cyrl & F+B & 22 &  &  \\
Shona & sna & shon1251 & Atlantic-Congo & Latn & F+B & 22 & \checkmark &  \\
Siamou & sif & siam1242 & Isolate & Latn & B & 19 & \checkmark & \checkmark \\
Sicilian & scn & sici1248 & Indo-European & Latn & F+B & 21 & \checkmark &  \\
Slovenian & slv & slov1268 & Indo-European & Latn & F+B & 22 & \checkmark &  \\
Somali & som & soma1255 & Afro-Asiatic & Latn & F+B & 20 & \checkmark &  \\
Southern Pashto & pbt & sout2649 & Indo-European & Arab & F+B & 22 &  &  \\
Southern Sotho & sot & sout2807 & Atlantic-Congo & Latn & F+B & 22 & \checkmark &  \\
Standard Arabic & arb & stan1318 & Afro-Asiatic & Arab & F & 22 &  &  \\
Standard Estonian & ekk & esto1258 & Uralic & Latn & F+B & 22 & \checkmark &  \\
Standard Latvian & lvs & latv1249 & Indo-European & Latn & F+B & 22 & \checkmark &  \\
Standard Malay & zsm & stan1306 & Austronesian & Latn & F+B & 21 & \checkmark &  \\
Sundanese & sun & sund1252 & Austronesian & Latn & F+B & 19 & \checkmark &  \\
Swahili & swh & swah1253 & Atlantic-Congo & Latn & F+B & 19 & \checkmark &  \\
Swedish & swe & swed1254 & Indo-European & Latn & F+B & 20 & \checkmark &  \\
Tagalog & tgl & taga1270 & Austronesian & Latn & B & 20 & \checkmark &  \\
Tamasheq & taq & tama1365 & Afro-Asiatic & Latn & F+B & 22 & \checkmark &  \\
Tamil & tam & tami1289 & Dravidian & Taml & F+B & 22 &  &  \\
Tatar & tat & tata1255 & Turkic & Cyrl & F+B & 21 &  &  \\
Tausug & tsg & taus1251 & Austronesian & Latn & B & 20 & \checkmark &  \\
Telugu & tel & telu1262 & Dravidian & Telu & F+B & 20 &  &  \\
Thai & tha & thai1261 & Tai-Kadai & Thai & F+B & 20 &  &  \\
Tibetan & bod & tibe1272 & Sino-Tibetan & Tibt & F+B & 22 &  &  \\
Tok Pisin & tpi & tokp1240 & Indo-European & Latn & F+B & 20 & \checkmark &  \\
Tosk Albanian & als & tosk1239 & Indo-European & Latn & F+B & 22 & \checkmark &  \\
Tsonga & tso & tson1249 & Atlantic-Congo & Latn & F+B & 20 & \checkmark &  \\
Tswana & tsn & tswa1253 & Atlantic-Congo & Latn & F+B & 22 & \checkmark &  \\
Turkish & tur & nucl1301 & Turkic & Latn & F+B & 22 & \checkmark &  \\
Turkmen & tuk & turk1304 & Turkic & Latn & F & 22 & \checkmark &  \\
Tuvinian & tyv & tuvi1240 & Turkic & Cyrl & F & 19 &  &  \\
Ukrainian & ukr & ukra1253 & Indo-European & Cyrl & F+B & 22 &  &  \\
Umbundu & umb & umbu1257 & Atlantic-Congo & Latn & F+B & 22 & \checkmark &  \\
Vietnamese & vie & viet1252 & Austroasiatic & Latn & F+B & 19 &  &  \\
Waray (Philippines) & war & wara1300 & Austronesian & Latn & F+B & 18 & \checkmark &  \\
Wayuu & guc & wayu1243 & Arawakan & Latn & B & 20 & \checkmark &  \\
Welsh & cym & wels1247 & Indo-European & Latn & F+B & 22 & \checkmark &  \\
Western Cham & cja & west2650 & Austronesian & Arab & B & 21 &  & \checkmark \\
Western Frisian & fry & west2354 & Indo-European & Latn & B & 22 & \checkmark &  \\
Wolof & wol & nucl1347 & Atlantic-Congo & Latn & F+B & 22 & \checkmark &  \\
Wu Chinese & wuu & wuch1236 & Sino-Tibetan & Hans & F+B & 22 &  &  \\
Xhosa & xho & xhos1239 & Atlantic-Congo & Latn & F+B & 21 & \checkmark &  \\
Xârâcùù & ane & xara1244 & Austronesian & Latn & B & 22 & \checkmark & \checkmark \\
Yemsa & jnj & yems1235 & Ta-Ne-Omotic & Latn & B & 21 & \checkmark & \checkmark \\
Yucateco & yua & yuca1254 & Mayan & Latn & B & 22 & \checkmark &  \\
Yue Chinese & yue & yuec1235 & Sino-Tibetan & Hant & F+B & 22 &  &  \\
Zande & zne & zand1248 & Atlantic-Congo & Latn & B & 21 & \checkmark &  \\
Zarma & dje & zarm1239 & Songhay & Latn & B & 22 & \checkmark &  \\
Zulu & zul & zulu1248 & Atlantic-Congo & Latn & F+B & 16 & \checkmark &  \\
\bottomrule
\end{tabular}
\caption{All 195 target languages (continued).}
\end{table*}

\begin{table*}[h]
\centering\footnotesize
\begin{tabular}{@{}>{\raggedright\arraybackslash}p{1.7cm}l>{\raggedright\arraybackslash}p{5.6cm}>{\raggedright\arraybackslash}p{3.2cm}rl@{}}
\toprule
Group & ID & Grambank question & Values & Items & Gold \\
\midrule
Word order & GB024 & What is the order of numeral and noun in the NP? & [1] Num--N; [2] N--Num; [3] both & 186 & 117/49/20 \\
 & GB025 & What is the order of adnominal demonstrative and noun? & [1] Dem--N; [2] N--Dem; [3] both & 186 & 102/58/26 \\
 & GB065 & What is the pragmatically unmarked order of adnominal possessor noun and possessed noun? & [1] Possessor--Possessed; [2] Possessed--Possessor; [3] both & 190 & 75/75/40 \\
\addlinespace[3pt]
 & GB130 & What is the pragmatically unmarked order of S and V in intransitive clauses? & [1] SV; [2] VS; [3] both & 194 & 157/26/11 \\
 & GB131 & Is a pragmatically unmarked constituent order verb-initial for transitive clauses? & [0] absent; [1] present & 191 & 159/32 \\
 & GB132 & Is a pragmatically unmarked constituent order verb-medial for transitive clauses? & [0] absent; [1] present & 191 & 90/101 \\
 & GB133 & Is a pragmatically unmarked constituent order verb-final for transitive clauses? & [0] absent; [1] present & 191 & 119/72 \\
 & GB193 & What is the order of adnominal property word and noun? & [0] they cannot be used attributively; [1] ANM--N; [2] N--ANM; [3] both & 181 & 0/81/75/25 \\
\midrule
Word classes & GB020 & Are there definite or specific articles? & [0] absent; [1] present & 186 & 103/83 \\
 & GB038 & Are there demonstrative classifiers? & [0] absent; [1] present & 186 & 180/6 \\
\addlinespace[3pt]
 & GB074 & Are there prepositions? & [0] absent; [1] present & 184 & 61/123 \\
 & GB075 & Are there postpositions? & [0] absent; [1] present & 181 & 95/86 \\
\addlinespace[3pt]
 & GB126 & Is there an existential verb? & [0] absent; [1] present & 162 & 56/106 \\
\midrule
Morphology & GB030 & Is there a gender distinction in independent 3rd person pronouns? & [0] absent; [1] present & 193 & 117/76 \\
 & GB042 & Is there productive overt morphological singular marking on nouns? & [0] absent; [1] present & 191 & 130/61 \\
 & GB044 & Is there productive morphological plural marking on nouns? & [0] absent; [1] present & 190 & 44/146 \\
\addlinespace[3pt]
 & GB082 & Is there overt morphological marking of present tense on verbs? & [0] absent; [1] present & 186 & 88/98 \\
 & GB083 & Is there overt morphological marking on the verb dedicated to past tense? & [0] absent; [1] present & 190 & 49/141 \\
 & GB084 & Is there overt morphological marking on the verb dedicated to future tense? & [0] absent; [1] present & 187 & 82/105 \\
\addlinespace[3pt]
 & GB137 & Can standard negation be marked clause-finally? & [0] absent; [1] present & 180 & 124/56 \\
 & GB138 & Can standard negation be marked clause-initially? & [0] absent; [1] present & 179 & 106/73 \\
\addlinespace[3pt]
 & GB172 & Can an article agree with the noun in gender/noun class? & [0] absent; [1] present & 179 & 147/32 \\
\bottomrule
\end{tabular}
\caption{The 22 core features. Items is the number of the 195 languages Grambank codes for the feature; Gold is how many of them have each value, in the order listed.}
\label{tab:core-features}
\end{table*}

\end{document}